\documentclass[10pt,journal]{IEEEtran}
\usepackage{cite}
\usepackage{multirow}  
\usepackage{listings}  
\usepackage{xcolor}
\usepackage{mathtools}
\usepackage[hyphens]{url}
\usepackage{graphicx}
\usepackage[ruled]{algorithm2e}
\usepackage{indentfirst}
\usepackage{graphicx}
\usepackage{epstopdf}
\usepackage{booktabs}
\usepackage{colortbl}
\usepackage{rotating}
\usepackage{multirow}
\usepackage{subfigure}
\usepackage{amsmath}
\usepackage{chngpage}
\usepackage{url}

\usepackage{mathrsfs}  

\begin{document}
\bibliographystyle{IEEEtran}
\title{QSAR Classification Modeling for Bioactivity of Molecular Structure via SPL-Logsum}


\author{Liang-Yong Xia,
        Qing-Yong Wang 
\thanks{L.-Y. Xia, Q.-Y.Wang are with Macau University of Science and Technology, Macau, China, 999078. wang@cnu.ac.cn}
}

\maketitle
\thispagestyle{empty}

\begin{abstract}
Quantitative structure-activity relationship (QSAR) modelling is effective 'bridge' to search the reliable relationship related bioactivity to molecular structure. A QSAR classification model contains a lager number of redundant, noisy and irrelevant descriptors. To address this problem, various of methods have been proposed for descriptor selection. Generally, they can be grouped into three categories: filters, wrappers, and embedded methods. Regularization method is an important embedded technology, which can be used for continuous shrinkage and automatic descriptors selection. In recent years, the interest of researchers in the application of regularization techniques is increasing in descriptors selection , such as, logistic regression(LR) with $L_1$ penalty. In this paper, we proposed a novel descriptor selection method based on self-paced learning(SPL) with Logsum penalized LR for predicting the bioactivity of molecular structure. SPL inspired by the learning process of humans and animals that gradually learns from easy samples(smaller losses) to hard samples(bigger losses) samples into training and Logsum regularization has capacity to select few meaningful and significant molecular descriptors, respectively. Experimental results on simulation and three public  QSAR datasets show that our proposed SPL-Logsum method outperforms other commonly used sparse methods in terms of  classification performance and model interpretation.
\end{abstract}

\begin{IEEEkeywords}
 QSAR; bioactivity; descriptor selection; SPL; Logsum
\end{IEEEkeywords}



\section{Introduction}
Quantitative structure-activity relationship (QSAR) model is effective 'bridge' to search the reliable relationship between chemical structure and biological activities in the field of drug design and discovery \cite{katritzky2010quantitative}.
The chemical structures are represented by a larger number of different descriptors. In general, only a few descriptors associated with bioactivity are favor of the QSAR model. Therefore, descriptor selection plays an important role in the study of QSAR that can eliminates redundant, noisy, and  irrelevant descriptors \cite{shahlaei2013descriptor}.

In the recent years, various of methods have been proposed for descriptor selection. Generally, they can be grouped into three categories: filters, wrappers, and embedded methods \cite{guyon2005result}.

Filter methodsselect descriptors according to some statistic criterion such as T-test. These selected descriptors will be a part of classification that used to classify the compounds \cite{duch2006filter} \cite{yu2003feature}.  The drawback of filter method ignores the relationship between descriptors.

Wrapper methods utilize a subset of descriptors and train a model using them\cite{karegowda2010feature}. For example, forward selection adds the most important descriptors until the model is not statistically significant\cite{whitley2000unsupervised}. Backward elimination starts with the candidate descriptor and remove descriptors without some statistical descriptors\cite{mao2004orthogonal}. Particle swarm optimization has a series of initial random particles and then selects descriptors by updating the velocity and positions\cite{kennedy2011particle}. Genetic algorithm initials random particles and then uses the code, selection, exchange and mutation operations to select the subset of descriptors\cite{deb2002fast}\cite{hemmateenejad2003genetic}. However, these methods are usually computationally very expensive.

Another method for descriptor selection is embedded method that combines filter methods and wrapper methods\cite{lal2006embedded}. Regularization method \cite{scholkopf2002learning}is an important embedded technology, which can be used for continuous shrinkage and automatic descriptors selection. Various of regularization types have been proposed , such as, Logsum\cite{candes2008enhancing},$L_1$\cite{tibshirani1996regression}, $L_{EN}$ \cite{zou2005regularization}, $L_{1/2}$\cite{xu20101},SCAD \cite{fan2001variable}, $MCP$\cite{zhang2010nearly}, $HLR (L_{1/2}+L_2)$\cite{huang2016feature} and so on. Recently, the regularization has been used in QSRR\cite{daghir2015least},QSPR \cite{goodarzi2010qspr} and QSTR \cite{aalizadeh2017prediction} in the field of chemometrics. However, some individuals have focused their interest and attention on QSAR research. Besides, the interest of applying regularization technology to LR is increasing in the QSAR classification study. The LR  model is considered to be an effective discriminant method because it provides the prediction probability of class members. In order to slelect the small subset of descriptors that can be for QSAR model of interest, various of regularized LR model have been development. For instance, Shevade SK et al\cite{shevade2003simple}. proposed sparse LR model with $L_1$ to extract the key variables for classification problem. However, the limitation of $L_1$  is  resulting in a biased estimation for large coefficients. Therefore, its model selection is inconsistent and lack of oracle properties. In order to alleviate these, Xu et al\cite{xu20101}.proposed $L_{1/2}$ penalty, which has demonstrated many attractive properties, such as unbiasedness, sparsity and oracle properties. Then, Liang et al.\cite{liang2013sparse} used LR based on $L_{1/2}$ penalty to select a small subset of variables.

\begin{figure*}[h]
    \centering
    \includegraphics[width=16cm]{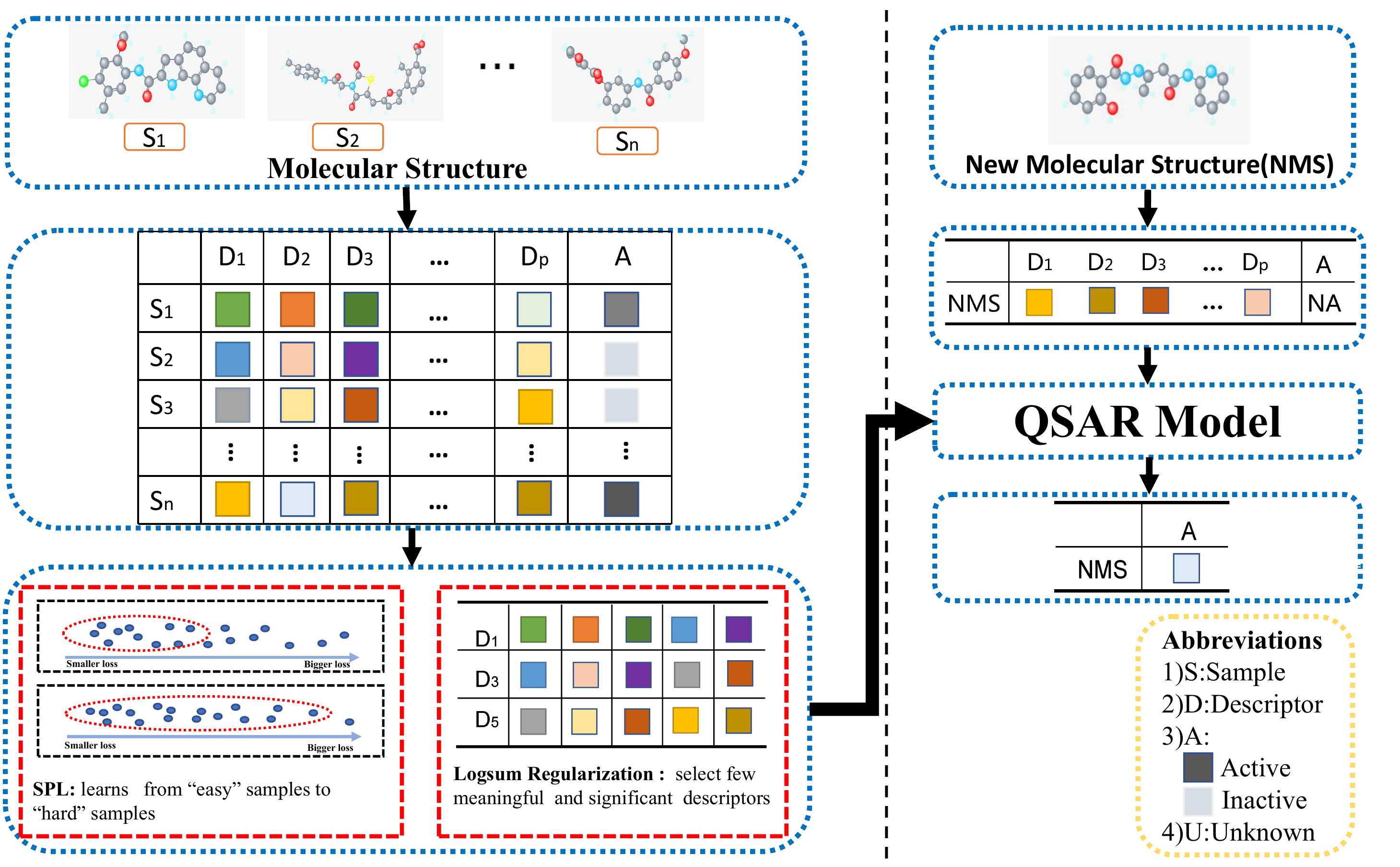}
    \caption{The diagram illustrates  the process of our proposed SPL-Logsum for QSAR modeling. The whole process is predicting molecular structure with unknown bioactivity which can be divided into five main stages.
    (1) Collecting molecular structure and biological actives; (2) Calculating molecular descriptors used by software QSARINS; (3) Learning and selecting from easy ones to hard ones and significant descriptors utilized by SPL and Logsum regularization, respectively; (4) Building the model with the optimum descriptor subset; (5) Predicting the bioactivity of a new molecular structure using the established model. Note that:different color blocks represent different values}
    \label{figure5}
\end{figure*}

In this paper, we proposed a novel descriptor selection using SPL via sparse LR with Logsum  penalty(SPL-Logsum) in QSAR classification. SPL is inspired by the learning process of humans that gradually incorporates the training samples into learning from easy ones to hard ones\cite{kumar2010self}\cite{jiang2014self}\cite{meng2015objective}. Different from the curriculum learning \cite{bengio2009curriculum} that learns the data in a predefined order based on prior knowledge, SPL learns the training data in an order from easy to hard dynamically determined by the feedback of the learner itself. Meanwhile, the Logsum regularization proposed by Candes et al\cite{candes2008enhancing}. produces better solutions with more sparsity. The flow diagram shows the process of our proposed SPL-Logsum for QSAR model in Fig. \ref{figure5}.

Our work has three main contributions:
\begin{itemize}
\item We integrate the self-paced learning into the Logsum penalized logistic regression\emph{(SPL-Logsum)}. Our proposed SPL-Logsum method can identify the easy and hard samples adaptively according to what the model has already learned and gradually add harder samples into training and prevent over-fitting simultaneously.
\item In the unbalanced data, our proposed method can still get good performance and be superior to other commonly used sparse methods.
\item Experimental results on both simulation and real datset corroborate our ideas and demonstrate the correctness and effectiveness of SPL-Logsum.
\end{itemize}

The structure of the paper is as follows: Section 2  briefly introduces Logsum penalized LR, SPL and classification evaluation criteria. The details of three QSAR datasets are given in Section 3. Section 4 provides the related experimental results on the artificial and three QSAR datasets. A concluding remark is finally made.

\section{Methods}
\subsection{\textbf{Sparse LR with the Logsum penalty}}
\subsubsection{\textbf{the Logsum penalized LR}}
In this paper, we only consider a general binary classification problem and  get a predictor vector X and a response variable y, which consists of chemical structure and corresponding biological activities, respectively. Suppose we have $n$ samples, $D$= {($X_1$, $y_1$), ($X_2$, $y_2$),..., ($X_n$, $y_n$)}, where $X_i$ = ($x_{i1}$, $x_{i2}$ ,..., $x_{ip}$) is $i^{th}$ input pattern with dimensionality $p$, which means the $X_i$ has $p$ descriptors and $x_{ij}$ denotes the value of descriptor j for the $i^{th}$ sample. And $y_i$ is a corresponding variable that takes a value of 0 or 1. Define a classifier $f(x)=e^x/(1+e^x)$ and the LR is given as follows:

\begin{equation}\label{1}
  P(y_i=1|X_i)=f(X_i^{'}\beta)=\frac{exp(X_i^{'})}{1+exp(X_i^{'}\beta)},
\end{equation}

where $\beta=(\beta_{0},\beta_{1},...,\beta_{p})$ are the coefficients  that can be to estimated , note that $\beta_{0}$ is the intercept. Additionally, the log-likelihood can be expressed as follows:

\begin{equation}\label{2}
  l(\beta)=-\sum_{i=1}^n\{y_{i}log[f(x^{'}_i\beta)]+(1-y_i)log[1-f(x^{'}_i\beta)]\},
\end{equation}

the $\beta$ can be got by minimizing the equation (\ref{2}). However, in high dimensional QSAR classification problem with $n>>p$, direct to solve the equation (\ref{2}) can result in over-fitting. Therefore, in order to solve the problem, add the regularization terms to the equation (\ref{2}).

\begin{equation}\label{3}
  \beta=argmin\{l(\beta)+\lambda p(\beta)\},
\end{equation}

where $l(\beta)$ is loss function, $p(\beta)$ is penalty function, $\lambda >0$ is a tuning parameter. Note that $p(\beta)=\sum|\beta|^{q}$. When $q$ is equal to 1, the $L_1$  has been proposed. Moreover, there are various of  versions of $L_1$, such as $L_{EN}$, SCAD, MCP, group lasso,and so on. We add the $L_1$ regularization to the equation (\ref{2}). The formula is expressed as follows:
\begin{equation}\label{4}
  \beta=argmin\{l(\beta)+\lambda\sum_{j=1}^{p}|\beta|\}
\end{equation}
$L_1$ regularization has capacity to select the descriptors. However, the drawback of $L_1$ regularization can result in a biased estimation for large coefficients and be not sparsity. To address these problems, Xu et al\cite{xu20101}. proposed $L_{1/2}$ regulation, which can be taken as a representative of $L_q$ $(0 < q < 1)$ penalty. We can rewrite the equation (\ref{4}) as follows:
\begin{equation}\label{5}
  \beta=argmin\{l(\beta)+\lambda\sum_{j=1}^{p}|\beta|^{\frac{1}{2}}\}
\end{equation}
Theoretically, the $L_0$ regularization produces better solutions with more sparsity , but it is an NP problem. Therefore, Candes et al\cite{candes2008enhancing}. proposed the Logsum penalty, which approximates the $L_0$ regularization much better. We give the formula based on Logsum regularization as follows:
\begin{equation}\label{6}
\beta=argmin\{l(\beta)+\lambda\sum_j^p log(|\beta_j|+\varepsilon)\}
\end{equation}
where $\varepsilon >0$ should be set arbitrarily small, to closely make the Logsum penalty resemble the $L_0$-norm.
\begin{figure}[htpb]
    \centering
    \includegraphics[width=9cm]{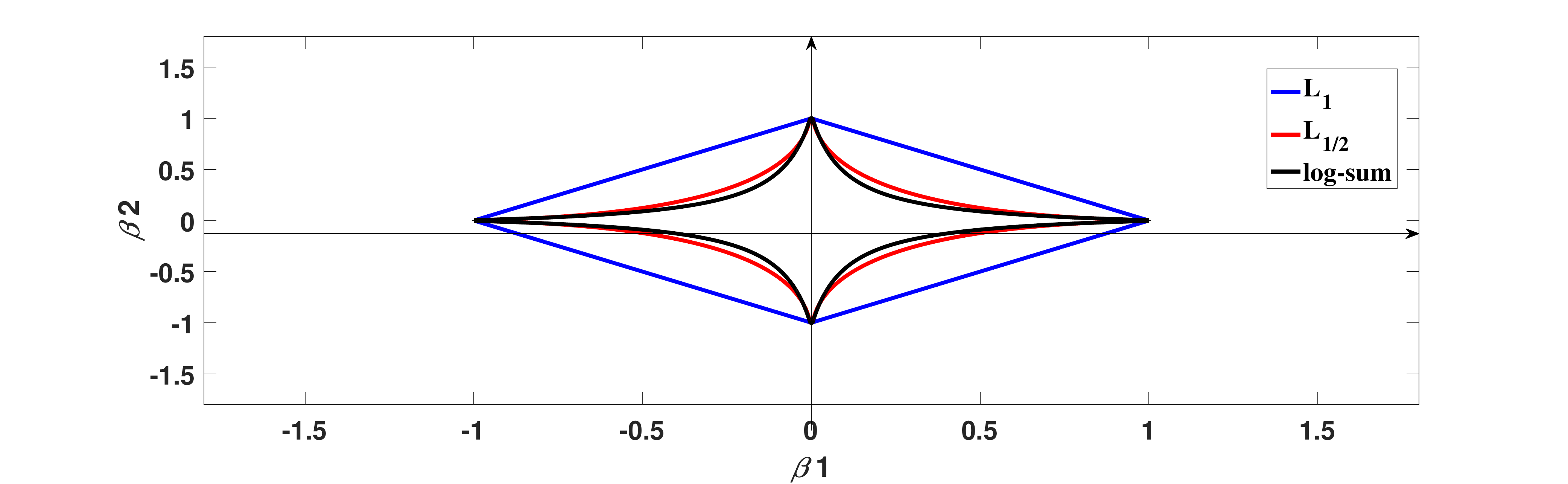}
    \caption{$L_1$ is convex,  $L_{1/2}$ and Logsum are non-convex. The Logsum approximates to $L_0$.}
    \label{convex}
\end{figure}
The Logsum regularization is non-convex in Fig. \ref{convex}.The Equation (\ref{6}) has a local minima \cite{xia2017descriptor}. A regularization can should satisfy three properties for the coefficient estimators: unbiasedness, sparsity and continuity in Fig. \ref{figure2}.
\begin{figure}[h]
    \centering
    \includegraphics[width=9cm]{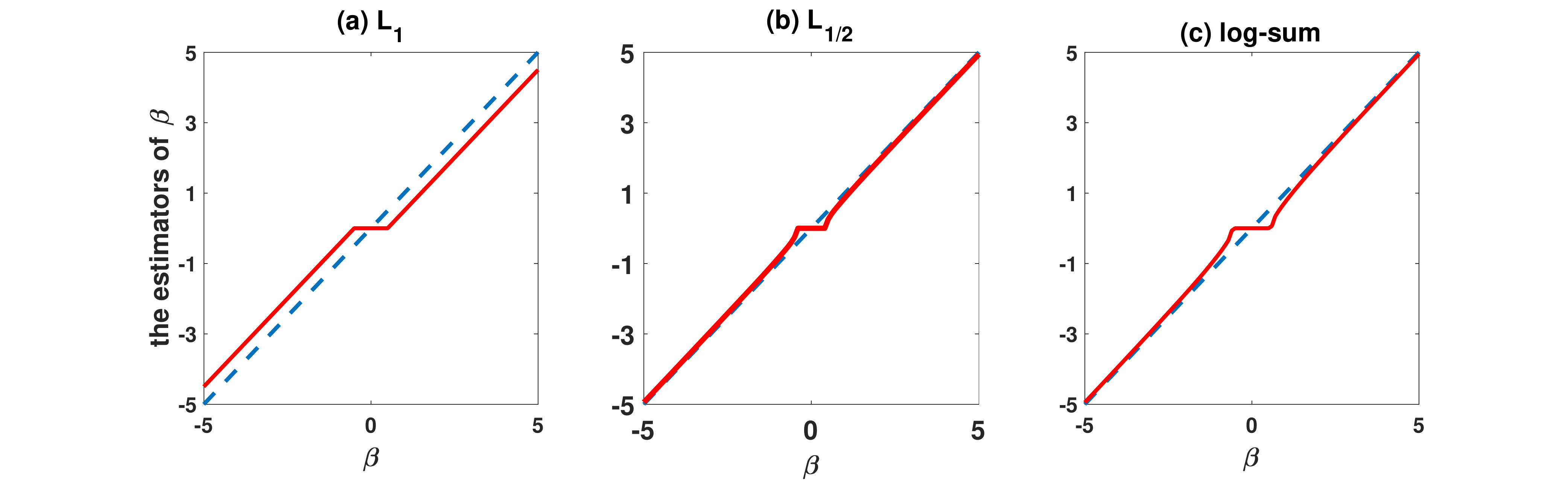}
    \caption{Three properties for the coefficient estimators: unbiasedness, sparsity and continuity for (a)$L_1$, (b)$L_{1/2}$ and (c) Logsum}
    \label{figure2}
\end{figure}

\subsubsection{\textbf{A coordinate descent algorithm for the Logsum penalized LR}}
In this paper, we used the coordinate descent algorithm to solve equation (\ref{6}). The algorithm is a "one-at-a-time" and solves $\beta_{j}$ and other $\beta_{j\neq k}$ (represent the parameters remained after $j^{th}$ element is removed) are fixed.
Suppose we have $n$ samples, $D$= {($X_1$, $y_1$), ($X_2$, $y_2$),..., ($X_n$, $y_n$)}, where $X_i$ = ($x_{i1}$, $x_{i2}$ ,..., $x_{ip}$) is $i^{th}$ input pattern with dimensionality $p$, which means the $X_i$ has $p$ descriptors and $x_{ij}$ denotes the value of descriptor $j$ for the $i^{th}$ sample. And $y_i$ is a corresponding variable that takes a value of 0 or 1.
According to Friedman et al. \cite{friedman2010regularization}, Liang et al. \cite{liang2013sparse} and Xia et al. \cite{xia2017descriptor}, the univariate Logsum thresholding operator can be written as:

\begin{equation}
\label{7}
\beta_j=D(w_j,\lambda,\varepsilon)=
\begin{cases}
   sign(w_j)\frac{c_1+\sqrt c_2}{2} &\mbox{if $c_2>0$ }\\
 0&\mbox{if $c_2\leq0 $}
   \end{cases}
\end{equation}

where $ \lambda>0,0<\varepsilon<\sqrt\lambda,c_1=\omega_j-\varepsilon$, $c_2=c_1^2-4(\lambda-w_j\varepsilon)$ and $w_j=\sum_{i=1}^{n}x_{ij}( y_{i}-\widetilde y_{i}^{(j)})$, $\widetilde y_{i}^{(j)}=\sum_{k\neq j}x_{ik}\beta_{k}$
\\
Inspired by Liang et al\cite{liang2013sparse}., the equation (\ref{6}) is linearized by one-term Taylor series expansion:
\begin{equation}\label{8}
  L(\beta,\lambda)\approx\frac{1}{2n}\sum_{i=1}^{n}
(Z_{i}-X_{i}\beta)^{'}W_{i}(Z_{i}-X_{i}\beta)+\lambda\sum_{j=1}^{n} log(|\beta_{j}|+\varepsilon)
\end{equation}

where $\varepsilon>0$,$Z_{i}=X_{i}\widetilde \beta+\frac{Y_{i}-f(X_{i}\widetilde\beta)}{f(X_{i}\widetilde\beta)(1-f(X_{i}\widetilde\beta))}$, $W_{i}=f(X_{i}\widetilde\beta)(1-f(X_{i}\widetilde\beta))$ and $f(X_{i}\widetilde\beta)=\frac{exp(X_{i}\widetilde\beta)}{(1+exp(X_{i}\widetilde\beta))}$. Redefine the partial residual for fitting $\widetilde\beta_{j}$ as $\widetilde Z_{i}^{(j)}=\sum_{i=1}^{n}W_{i}(\widetilde Z_{i}-\sum_{k\neq j}x_{ik}\widetilde\beta_{k})$ and $\sum_{i=1}^{n}x_{ij}(Z_{i}-\widetilde Z_{i}^{(j)})$. A  pseudocode of  coordinate descent algorithm for Logsum penalized LR model is shown in Algorithm \ref{alg1}.
\\

\begin{algorithm}
\caption{A coordinate descent algorithm for Logsum penalized LR model}
\label{alg1}
\KwIn{X,y and $\lambda$  is chosen by 10-fold cross-validation}
\KwOut{$\beta$ and the value of loss}
\While{$\beta(m)$dose convergence } {
Initialize all $\beta_j(m)=0 (j=1,2,3,...,p),\varepsilon$,set $m=0$\\
Calculate $Z(m)$ and $W(m)$ and the loss function E.q. (\ref{8}) based on $\beta(m)$\\
Update each $\beta_j(m)$and cycle $j=1,2,3,...,p$\\
\hspace*{1.5cm}
$\widetilde z_{i}^{(j)}(m)\leftarrow \sum_{k\neq j}x_{ik}\beta_{k}(m)$\\
\hspace*{2cm}and $w_j(m) \leftarrow w_{j}(m)x_{ij}(Z_{i}(m)-\widetilde Z_{i}^{(j)}(m))$\\
Update $\beta_j(m)=D(w_j,\lambda,\varepsilon)$\\
Let $m \leftarrow (m+1)$,$\beta(m+1)\leftarrow \beta(m)$\\
}
\end{algorithm}


\subsection{\textbf{SPL-Logsum}}
Inspired by cognitive mechanism of humans and animals, Koller et al\cite{kumar2010self}. proposed a new learning called SPL (SPL) that learns from easy to hard samples. During the process of optimization, more samples are entered into training set from easy to hard by increasing gradually the penalty of the SPL regularizer. Suppose given a dataset $D={(X_{i},y_{i})}_{i=1}^{n}$ with $n$ samples. $X_{i}$ and $y_{i}$ are the $i^{th}$ sample and its label,respectively. The value of $y_{i}$ is 0 or 1 in classification model. Let $f(X_{i},\beta)$ denote the learn model and $\beta$ is model parmaeter that should be estimated. $l(y_{i},f(X_{i},\beta))$ is loss function of the $i^{th}$ sample. The SPL model combines a weighted loss term and a general self-paced regularizer imposed on sample weight, given as:

\begin{equation}
\label{9}
\begin{aligned}
 min_{\beta,v\in[0,1]^{n}} E(\beta,v,\gamma,\lambda) = \sum_{i=1}^{n}(v_{i}l(y_{i},f(X_{i},\beta))    \\
  +\lambda p(\beta)+g(v_{i},\gamma))
\end{aligned}
\end{equation}

where $g(v_{i},\gamma)=-\gamma v_{i}$ and $p(\beta)=log(|\beta|+\varepsilon)$. Therefore, the equation (\ref{9}) can be rewritten as:

\begin{equation}\label{10}
\begin{aligned}
min_{\beta,v\in[0,1]^{n}} E(\beta,v,\gamma,\lambda)=\sum_{i=1}^{n}(v_{i}l(y_{i},f(X_{i},\beta))  \\
 +\lambda log(|\beta|+\varepsilon)-\gamma v_{i})
\end{aligned}
\end{equation}
where $\lambda $ is regularization parameter , $\gamma$ is age parameter for controlling the learning pace,and $\varepsilon >0$. According to Kumar et al\cite{kumar2010self}, with the fixed $\beta$, the optimal $v_{i}^{*}$ is easily calculated by:

\begin{equation}\label{11}
  v_{i}^{*}=\begin{cases}
   1 &\mbox{if $l(y_{i},f(X_{i},\beta))<\gamma$ }\\
 0&\mbox{otherwise}
   \end{cases}
\end{equation}

\begin{figure}[htpb]
    \centering
    \includegraphics[width=8cm]{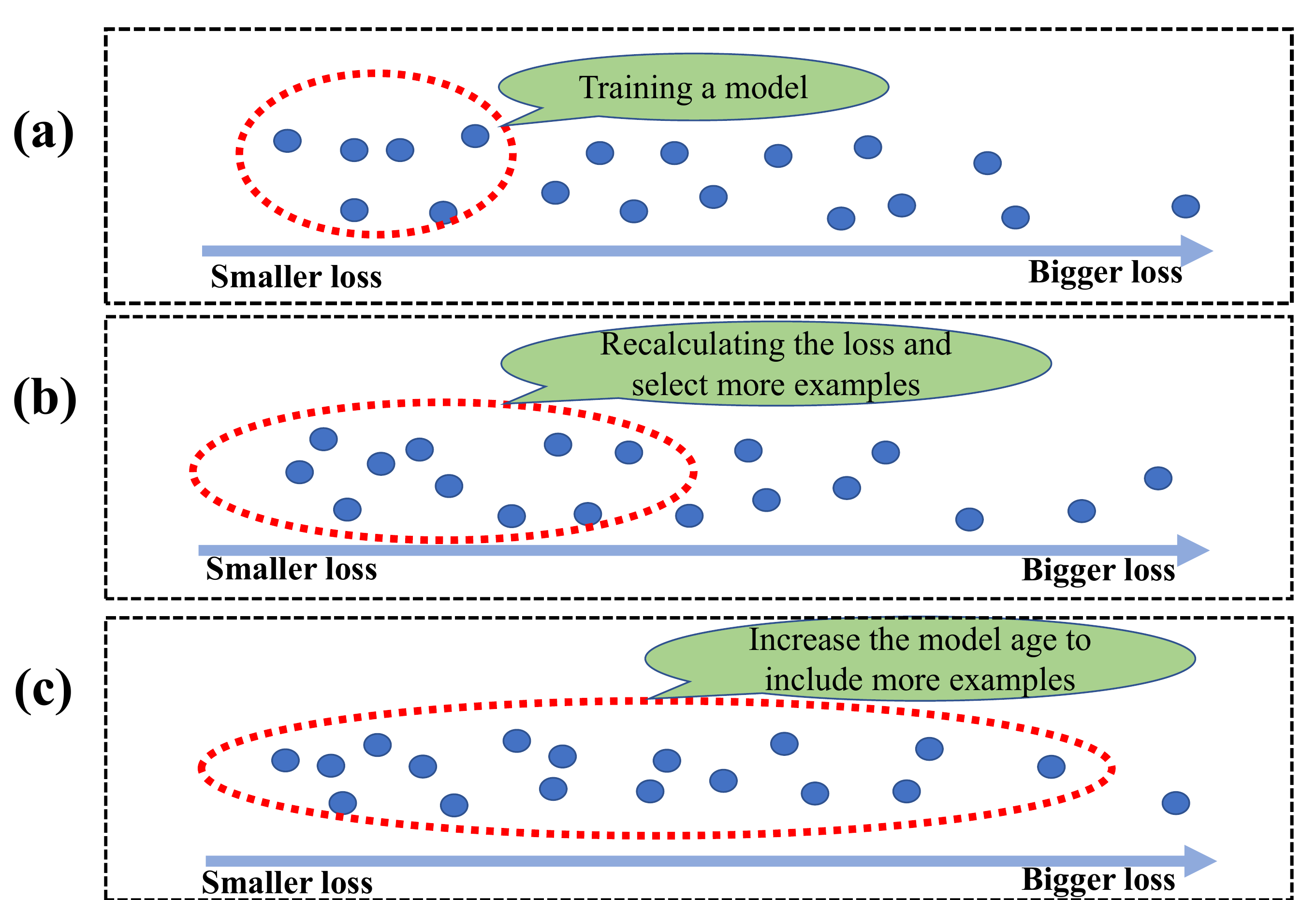}
    \caption{The process of our proposed SPL-Logsum is for selecting sampls:(a)Fixed $v$, and optimal model parameter $\beta$  (b)Fixed $\beta$, and optimal weight  $v$ (c)Increase the model age $\gamma$ to train more hard model}
    \label{figure3}
\end{figure}

Here, we give a explanation for the equation (\ref{11}). In SPL iteration,when estimating  the $v$ with a fixed $\beta$, a sample, which is taken as an easy sample (high-confidence sample with smaller loss value), can be selected($v^{*}=1$) in training if the loss is smaller than $\gamma$. Otherwise, unselected ($v^{*}=0$). When estimating the $\beta$ with a fixed $v$, the classifier only is used easy samples as training set. As the model "age" of $\gamma$ is increasing,  it means that more, probably hard samples with larger losses  will be considered to train a more "mature" model. The process of our proposed SPL-Logsum method is shown in Fig. \ref{figure3}. Besides, A  pseudocode of our proposed SPL-Logsum is shown in Algorithm \ref{alg2}.

\begin{algorithm}
\caption{Pseudocode for SPL-Logsum }
\label{alg2}
\KwIn{Input X,y,a stepsize $\mu$ and the "age" $\gamma$ }
\KwOut{Model parameter $\beta$}
Initialize $v^{*}$ and the value of loss \\
\While{not converged do } {
\hspace*{1cm}   update $\beta^{*}$  used by Algorithm I\\
 \hspace*{1cm}   update $v^{*}$ according to E.q. (\ref{11})\\
 \hspace*{0.7cm}increase $\gamma$ by the stepsize $\mu$\\
}
 \textbf{return }$\beta^{*}$\\
\end{algorithm}

%

\subsection{\textbf{Classification evaluation criteria}}
In order to evaluate the QSAR classification performance of proposed method, five classification evaluation criteria are implemented:(1)Accuracy,(2)Area under the curve (AUC), (3)Sensitivity, (4)Specificity and (5)P-value, which whether the selected descriptors are significant. In order to evaluate the performance of descriptors selection in simulation data, the sensitivity and specificity are defined as follows\cite{huang2016feature}\cite{zhang2013molecular}:\\

True Negative(TN)$:=|\bar{\beta}{.}*\bar{\hat{\beta}}|_{0}$, False Negative(FN)$:=|\beta{.}*\bar{\hat{\beta}}|_{0}$ \\

True Positive(TP)$:=|\beta{.}*\hat{\beta}|_{0}$, False Positive(FP)$:=|\bar{\beta}{.}*\hat{\beta}|_{0}$ \\

\hspace*{0.7cm}Sensitivity$:=\frac{TP}{TP+FN}$,specificity $:=\frac{TN}{TN+FP}$
\\where $.*$ is the element-wise product. $|.|_{0}$ calculates the number of non-zero element in a vector. are the on the vectors. The logical"not" operators of $\beta$ and $\hat{\beta}$ is $\bar{\beta}$ and $\bar{\hat{\beta}}$, respectively.
\section{Datasets}
\subsection{\textbf{PubChem AID:651580}}
The dataset is provided by the Anderson cancer center at the university of Texas. There are 1,759 samples, of which 982 is active, and the 777 is inactive. The 1,875 descriptors can be calculated using QSARINS software. After pretreatment, we used 1,614 samples, of which 914 is active and 700 is inactive. Each sample contains 1,642 descriptors \cite{651580}.
\subsection{\textbf{PubChem AID:743297}}
We could get this dataset from website\cite{743297}. By using QSARINS software and preprocessing, we utilized 200 samples, which consist of 58 active  and 142 inactive, with 1588 descriptors for model as input.
\subsection{\textbf{PubChem AID:743263}}
The QSAR dataset is from Georgetown University. The number of active and inactive is 252 and 868, respectively. After preprocessing, 983 samples could be used for QSAR modelling. Each sample contains 1,612 descriptors. Note that know more about details from website\cite{743263}.

\section{Results}
\subsection{Analyses of simulated data}
In this section, we evaluate the performance of our proposed SPL-logsum method in the simulation study. Four methods are compared to our proposed method, including LR with $L_{1}$, $L_{1/2}$, and Logsum penalties, respectively. In addition, Some factors will be considered for constructing simulation, including the value of loss, model "age" $\gamma$, weight $v$, the confidence of sample, sample size $n$, correlation coefficient $\rho$ and  noise control parameter $\sigma$.

\subsubsection{Loss, "age" $\gamma$ and weight $v$}
 A sample, which is taken as an easy sample  with small loss value , can be selected($v^{*}=1$) in training if the loss is smaller than $\gamma$. Otherwise, unselected ($v^{*}=0$). As the model "age" of $\gamma$ is increasing, more probably hard samples with larger losses  will be considered to train a more "mature" model. Therefore, we constructed a simple simulation about loss, $\gamma$ and $v$. First of all, a group of the value of loss is given. Then the model "age" is pre-set. At last, the selected samples are based on equation ($\ref{11}$). Table \ref{loss} shows that the selection of samples is increasing with "age" $\gamma$ \cite{jiang2015self}.

 \begin{table}[h]
\footnotesize
  \centering
  \caption{In SPL iteration, when the model ``age" $\gamma$ is increasing, more samples are included to be trained. Probably, hard samples with larger losses will be considered to train a more ``mature" model.}
  \setlength{\tabcolsep}{0.45mm}{
    \begin{tabular}{|c|cccccccccccccc|}
    \toprule
    \multicolumn{1}{|l|}{\textbf{Samples}} & \textit{\textbf{A}} & \textit{\textbf{B}} & \textit{\textbf{C}} & \textit{\textbf{D}} & \textit{\textbf{E}} & \textit{\textbf{F}} & \textit{\textbf{G}} & \textit{\textbf{H}} & \textit{\textbf{I}} & \textit{\textbf{J}} & \textit{\textbf{K}} & \textit{\textbf{L}} & \textit{\textbf{M}} & \textit{\textbf{N}} \\
    \midrule
    \multicolumn{1}{|l|}{\textbf{Loss}} & \textbf{0.05} & \textbf{0.12} & \textbf{0.12} & \textbf{0.12} & \textbf{0.15} & \textbf{0.4} & \textbf{0.2} & \textbf{0.18} & \textbf{0.35} & \textbf{0.15} & \textbf{0.16} & \textbf{0.2} & \textbf{0.5} & \textbf{0.3} \\
    \midrule
    \multicolumn{15}{c}{\textit{\textbf{When the ``age" \textbf{$\gamma=0.15$}}}} \\
    \midrule
    \textit{\textbf{V}} & \textbf{1} & \textbf{1} & \textbf{1} & \textbf{1} & \textbf{0} & \textbf{0} & \textbf{0} & \textbf{0} & \textbf{0} & \textbf{0} & \textbf{0} & \textbf{0} & \textbf{0} & \textbf{0} \\
    \midrule
    \textit{\textbf{SPL selects:}} & \textit{\textbf{A}} & \textit{\textbf{B}} & \textit{\textbf{C}} & \textit{\textbf{D}} &       &       &       &       &       &       &       &       &       &  \\
    \midrule
    \multicolumn{15}{c}{\textit{\textbf{  When the ``age" \textbf{$\gamma=0.2$}}}} \\
    \midrule
    \textit{\textbf{SPL selects:}} & \textit{\textbf{A}} & \textit{\textbf{B}} & \textit{\textbf{C}} & \textit{\textbf{D}} & \textit{\textbf{E}} &       & \textit{\textbf{G}} & \textit{\textbf{H}} &       & \textit{\textbf{J}} & \textit{\textbf{K}} &       &       &  \\
    \midrule
    \multicolumn{15}{c}{\textit{\textbf{When the ``age" \textbf{$\gamma=0.25$}}}} \\
    \midrule
    \textit{\textbf{V}} & \textbf{1} & \textbf{1} & \textbf{1} & \textbf{1} & \textbf{1} & \textbf{0} & \textbf{1} & \textbf{1} & \textbf{0} & \textbf{1} & \textbf{1} & \textbf{1} & \textbf{0} & \textbf{0} \\
    \midrule
    \textit{\textbf{SPL selects:}} & \textit{\textbf{A}} & \textit{\textbf{B}} & \textit{\textbf{C}} & \textit{\textbf{D}} & \textit{\textbf{E}} &       & \textit{\textbf{G}} & \textit{\textbf{H}} &       & \textit{\textbf{J}} & \textit{\textbf{K}} & \textit{\textbf{L}} &       &  \\
    \midrule
    \multicolumn{15}{c}{\textit{\textbf{When the ``age" \textbf{$\gamma=0.3$}}}} \\
    \midrule
    \textit{\textbf{V}} & \textbf{1} & \textbf{1} & \textbf{1} & \textbf{1} & \textbf{1} & \textbf{0} & \textbf{1} & \textbf{1} & \textbf{0} & \textbf{1} & \textbf{1} & \textbf{1} & \textbf{0} & \textbf{1} \\
    \midrule
    \textit{\textbf{SPL selects:}} & \textit{\textbf{A}} & \textit{\textbf{B}} & \textit{\textbf{C}} & \textit{\textbf{D}} & \textit{\textbf{E}} &       & \textit{\textbf{G}} & \textit{\textbf{H}} &       & \textit{\textbf{J}} & \textit{\textbf{K}} & \textit{\textbf{L}} &       & \textit{\textbf{N}} \\
    \midrule
    \bottomrule
    \end{tabular}%
  \label{loss}}%
\end{table}%

\subsubsection{High-confidence samples, medium-confidence samples, low-confidence samples}
\begin{figure}[h]
    \centering
    \includegraphics[width=8cm]{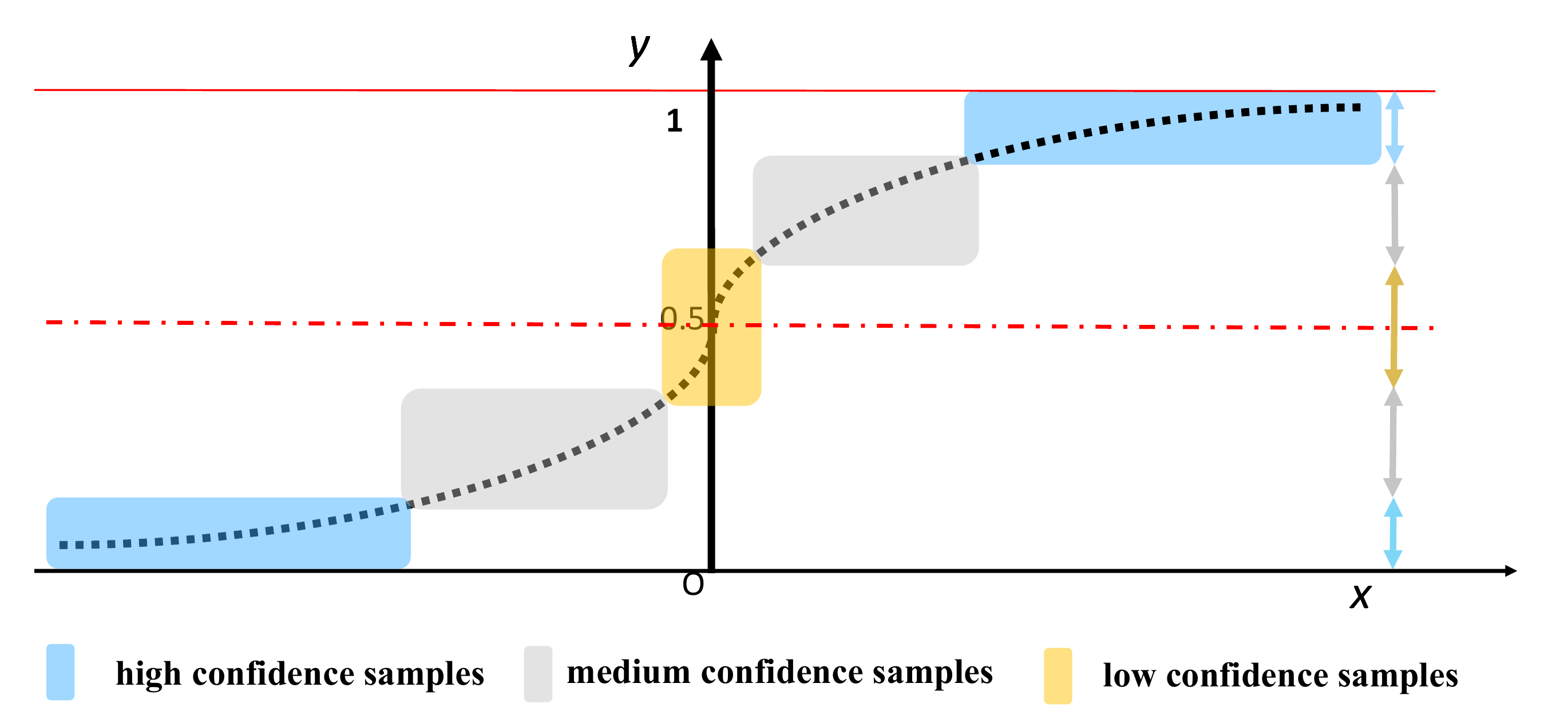}
    \caption{A set of samples can be divided into three parts. The color blue, gray and yellow blocks represent high-confidence samples, medium-confidence samples and low-confidence samples, respectively.}
    \label{figure6}
\end{figure}

\begin{figure}[htpb]
    \centering
    \includegraphics[width=8cm]{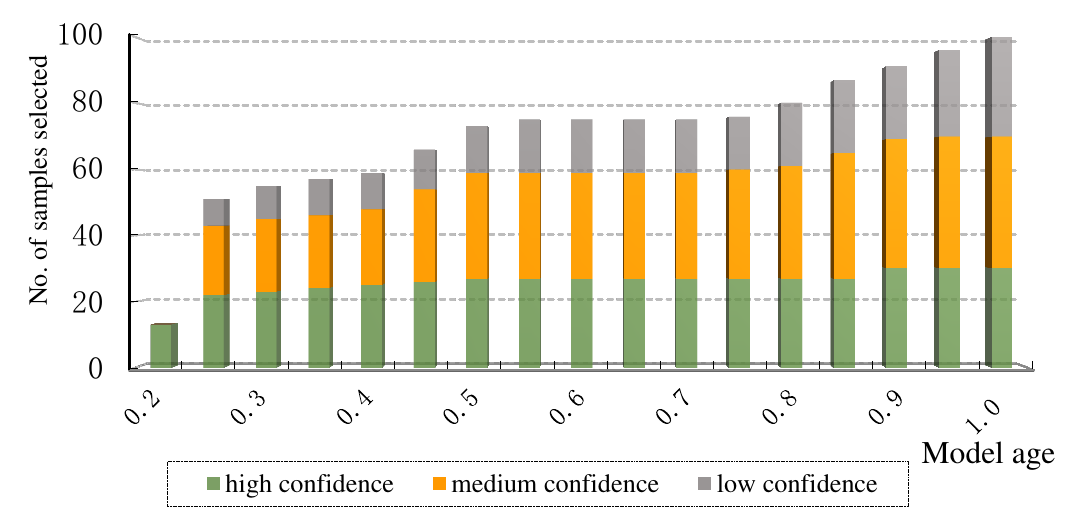}
    \caption{The results got by SPL. The number of samples is selected while the model age is increasing.($\gamma=0.2$ $\mu=0.05)$ }
    \label{figure7}
\end{figure}

In this paper, we divided samples into three parts, including high-confidence samples, medium-confidence samples and low-confidence samples. The high-confidence samples are favored by the model, followed by medium-confidence samples and low-confidence samples. The low-confidence samples are probably noise or outline that can reduce the performance of QSAR model. In order to illustrate the process of selected samples for SPL, we constructed a simple simulation. First of all, the simulated data was generated from the LR with using normal distribution to produce $X^{n \times p}$, where $n=100,p=1000$. Then, a set of coefficients $\beta$ is given. Finally, we can calculate the value of $y$. Fig.\ref{figure6}  shows that a set of samples can be divided into three parts. Therefore, in SPL iteration, we could count up the number of selected samples, where is from high-confidence samples, medium-confidence samples and low-confidence samples. Fig.\ref{figure7} indicates that at the beginning of the SPL, the model inclines to select high-confidence samples. Afterwards, when the model age gets larger, it tends to incorporate medium-confidence and low-confidence samples to train a mature model.


\subsubsection{Sample size $n$,correlation coefficient $\rho$ and  noise control parameter $\sigma$ }
In this section, we constructed a simulation with sample size $n=200,300$, correlation coefficient $\rho=0.2,0.6$ and  noise control parameter $\sigma=0.3,0.9$. The process of construction is expressed as:\\
Step I: use normal distribution to produce X. Here, the number of row(X) is sample $(n)$  and the number of column(X) is variable $(p=1000)$, respectively.
\begin{equation}\label{12}
  log(\frac{y_{i}}{1-y_{i}})=\beta_{0}+\sum_{j=1}^{p}x_{ij}\beta_{j}+\sigma\epsilon
\end{equation}
where $y=(y_1,......,y_n)^T$ is the vector of $n$ response variables, X = \{$X_1$,$X_2$,......,$X_n$\} is the generated matrix with $X_i=(x_{i1},......,x_{ip})$, $ \epsilon = (\epsilon_1,......,\epsilon_n)^T$ is the random error, $\sigma$ controls the signal to noise.\\
Step II: add different correlation parameter $\rho$ to simulation data.
\begin{equation}\label{13}
x_{ij}=\rho \times x_{11}+ (1-\rho)x_{ij} ,i\sim(1,......,n),j\sim(2,3,4,5)
\end{equation}
Step III: validate variable selection, the coefficients(10) are set in advance from 1 to 10.
\begin{equation}\label{14}
\beta=\overbrace{\underbrace{1,-1,-1.5,-3,2......,2,}_{10}\underbrace{0,0,0,......,0}_{990}}^{1000}
\end{equation}
Where $\beta$ is the coefficient.\\
Step IV: we can get y from equation $(\ref{12}),(\ref{13}), (\ref{14})$.\\

In  this simulation study, first of all, 100 groups of data are constructed with $n$, $\rho$ and $\sigma$. Then, divide the dataset into training set and testing set(training set:testing set=7:3 ). And then the coefficients are pre-set in advance. Finally, the LR with different penalties are used to select variables and builded model, including our proposed method. Note that the results should be averaged.

\begin{table}[h]
\scriptsize
  \centering
  \caption{The results are got by different methods with different $n$, $\rho$ and $\sigma$. }
   \setlength{\tabcolsep}{0.4mm}{
    \begin{tabular}{cccccccc|ccc}
    \toprule
    \multicolumn{3}{c|}{\textit{\textbf{Factors}}} & \multicolumn{1}{c|}{\multirow{2}[4]{*}{\textbf{Methods }}} & \multicolumn{4}{c|}{\textit{\textbf{Testing dataset}}} &       & \multicolumn{2}{c}{\textit{\textbf{$\beta$}}} \\
\cmidrule{1-3}\cmidrule{5-11}    \textbf{$n$} & \textbf{$\rho$} & \multicolumn{1}{c|}{\textbf{$\sigma$}} & \multicolumn{1}{c|}{} & \textit{\textbf{AUC}} & \textit{\textbf{Sensitivity}} & \textit{\textbf{Specificity}} & \textit{\textbf{Accuracy}} &       & \textit{\textbf{Sensitivity}} & \textit{\textbf{Specificity}} \\
\cmidrule{1-8}\cmidrule{10-11}    \multirow{16}[32]{*}{\textit{\textbf{200}}} & \multirow{8}[16]{*}{\textit{\textbf{0.2}}} & \multirow{4}[8]{*}{\textit{\textbf{0.3}}} & \textbf{$L_1$} & 0.7056 & 0.7000 & 0.7667 & 0.7333 &       & 0.8000 & 0.9495 \\
\cmidrule{4-4}          &       &       & \textbf{$L_{1/2}$} & 0.7133 & 0.7667 & 0.7333 & 0.7500 &       & 0.7000 & 0.9889 \\
\cmidrule{4-4}          &       &       & \textbf{Logsum} & 0.7578 & 0.8333 & 0.7667 & 0.8000 &       & 0.8000 & 0.9939 \\
\cmidrule{4-4}          &       &       & \textbf{SPL-Logsum} & \textit{\textbf{0.8711}} & \textit{\textbf{0.8333}} & \textit{\textbf{0.9000}} & \textit{\textbf{0.8667}} &       & \textit{\textbf{0.8000}} & \textit{\textbf{0.9980}} \\
\cmidrule{3-11}          &       & \multirow{4}[8]{*}{\textit{\textbf{0.9}}} & \textbf{$L_1$} & 0.6296 & 0.5862 & 0.6774 & 0.6333 &       & 0.7000 & 0.9283 \\
\cmidrule{4-4}          &       &       & \textbf{$L_{1/2}$} & 0.7397 & 0.6552 & 0.7742 & 0.7167 &       & 0.5000 & 0.9869 \\
\cmidrule{4-4}          &       &       & \textbf{Logsum} & 0.7553 & 0.6897 & 0.7742 & 0.7333 &       & 0.7000 & 0.9919 \\
\cmidrule{4-4}          &       &       & \textbf{SPL-Logsum} & \textit{\textbf{0.7998}} & \textit{\textbf{0.7241}} & \textit{\textbf{0.8710}} & \textit{\textbf{0.8000}} &       & \textit{\textbf{0.7000}} & \textit{\textbf{0.9970}} \\
\cmidrule{2-11}          & \multirow{8}[16]{*}{\textit{\textbf{0.6}}} & \multirow{4}[8]{*}{\textit{\textbf{0.3}}} & \textbf{$L_1$} & 0.8058 & 0.7576 & 0.8519 & 0.8000 &       & 0.6000 & 0.9556 \\
\cmidrule{4-4}          &       &       & \textbf{$L_{1/2}$} & 0.8586 & 0.8182 & 0.8889 & 0.8500 &       & 0.6000 & 0.9939 \\
\cmidrule{4-4}          &       &       & \textbf{Logsum} & 0.8260 & 0.8182 & 0.8148 & 0.8167 &       & 0.6000 & 0.9960 \\
\cmidrule{4-4}          &       &       & \textbf{SPL-Logsum} & \textit{\textbf{0.8709}} & \textit{\textbf{0.8485}} & \textit{\textbf{0.8889}} & \textit{\textbf{0.8667}} &       & \textit{\textbf{0.6000}} & \textit{\textbf{0.9970}} \\
\cmidrule{3-11}          &       & \multirow{4}[8]{*}{\textit{\textbf{0.9}}} & \textbf{$L_1$} & 0.8880 & 0.7429 & 0.9600 & 0.8333 &       & 0.6000 & 0.9495 \\
\cmidrule{4-4}          &       &       & \textbf{$L_{1/2}$} & 0.8480 & 0.7714 & 0.8800 & 0.8167 &       & 0.6000 & 0.9939 \\
\cmidrule{4-4}          &       &       & \textbf{Logsum} & 0.8903 & 0.8000 & 0.9600 & 0.8667 &       & 0.6000 & 0.9929 \\
\cmidrule{4-4}          &       &       & \textbf{SPL-Logsum} & \textit{\textbf{0.8960}} & \textit{\textbf{0.7714}} & \textit{\textbf{1.0000}} & \textit{\textbf{0.8667}} &       & \textit{\textbf{0.6000}} & \textit{\textbf{0.9970}} \\
    \midrule
    \multirow{16}[32]{*}{\textit{\textbf{300}}} & \multirow{8}[16]{*}{\textit{\textbf{0.2}}} & \multirow{4}[8]{*}{\textit{\textbf{0.3}}} & \textbf{$L_1$} & 0.9180 & 0.9111 & 0.9111 & 0.9111 &       & 1.0000 & 0.9323 \\
\cmidrule{4-4}          &       &       & \textbf{$L_{1/2}$} & 0.9536 & 0.9556 & 0.9333 & 0.9444 &       & 1.0000 & 0.9949 \\
\cmidrule{4-4}          &       &       & \textbf{logsum} & 0.9832 & 0.9556 & 0.9778 & 0.9667 &       & 1.0000 & 0.9980 \\
\cmidrule{4-4}          &       &       & \textbf{SPL-Logsum} & \textit{\textbf{0.9832}} & \textit{\textbf{0.9556}} & \textit{\textbf{0.9778}} & \textit{\textbf{0.9667}} &       & \textit{\textbf{1.0000}} & \textit{\textbf{0.9980}} \\
\cmidrule{3-11}          &       & \multirow{4}[8]{*}{\textit{\textbf{0.9}}} & \textbf{$L_1$} & 0.8392 & 0.7907 & 0.8085 & 0.8000 &       & 1.0000 & 0.9465 \\
\cmidrule{4-4}          &       &       & \textbf{$L_{1/2}$} & 0.8075 & 0.7674 & 0.7872 & 0.7778 &       & 0.9000 & 0.9869 \\
\cmidrule{4-4}          &       &       & \textbf{Logsum} & 0.8545 & 0.7907 & 0.8298 & 0.8111 &       & 0.9000 & 0.9899 \\
\cmidrule{4-4}          &       &       & \textbf{SPL-Logsum} & \textit{\textbf{0.9030}} & \textit{\textbf{0.9070}} & \textit{\textbf{0.8511}} & \textit{\textbf{0.8778}} &       & \textit{\textbf{1.0000}} & \textit{\textbf{0.9980}} \\
\cmidrule{2-11}          & \multirow{8}[16]{*}{\textit{\textbf{0.6}}} & \multirow{4}[8]{*}{\textit{\textbf{0.3}}} & \textbf{$L_1$} & 0.9071 & 0.8409 & 0.9348 & 0.8889 &       & 0.7000 & 0.9919 \\
\cmidrule{4-4}          &       &       & \textbf{$L_{1/2}$} & 0.9175 & 0.9091 & 0.9130 & 0.9111 &       & 0.6000 & 0.9960 \\
\cmidrule{4-4}          &       &       & \textbf{Logsum} & 0.9086 & 0.8409 & 0.9348 & 0.8889 &       & 0.6000 & 0.9929 \\
\cmidrule{4-4}          &       &       & \textbf{SPL-Logsum} & \textit{\textbf{0.9506}} & \textit{\textbf{0.9091}} & \textit{\textbf{0.9348}} & \textit{\textbf{0.9222}} &       & \textit{\textbf{0.7000}} & \textit{\textbf{0.9980}} \\
\cmidrule{3-11}          &       & \multirow{4}[8]{*}{\textit{\textbf{0.9}}} & \textbf{$L_1$} & 0.7319 & 0.8235 & 0.7115 & 0.7750 &       & 0.7000 & 0.8970 \\
\cmidrule{4-4}          &       &       & \textbf{$L_{1/2}$} & 0.7432 & 0.7941 & 0.7500 & 0.7750 &       & 0.6000 & 0.9758 \\
\cmidrule{4-4}          &       &       & \textbf{Logsum} & 0.8200 & 0.8250 & 0.7600 & 0.7889 &       & 0.6000 & 0.9879 \\
\cmidrule{4-4}          &       &       & \textbf{SPL-Logsum} & \textit{\textbf{0.9310}} & \textit{\textbf{0.8750}} & \textit{\textbf{0.9000}} & \textit{\textbf{0.8889}} &       & \textit{\textbf{0.7000}} & \textit{\textbf{0.9970}} \\
    \bottomrule
    \end{tabular}%
  \label{table2}}%
\end{table}%

According to existing literature \cite{golbraikh2002beware}, we have learned that the predictive ability of a QSAR model can only be estimated using testing set of compounds. Therefore, we poured more interest and attention into testing dataset, of which can prove the generalization ability of the model. Table \ref{table2} exhibits that the experiment results are got by $L_1$, $L_{1/2}$ and Logsum, including our proposed SPL-Logsum methods. The performance of $ \beta$  got by our proposed SPL-lgosum method are better than those of $L_1$, $L_{1/2}$ and Logsum. For example, when $n=200$,$\rho=0.2$ and $\sigma=0.3$, the sensitivity of $L_1$, Logsum and SPL-Logsum is 0.8, 0.8 and 0.8 higher than 0.7 of $L_{1/2}$. the specificity obtained by our proposed SPL-Logsum is the highest among four methods with 0.9980. Logsum , $L_{1/2}$, $L_1$ rank in the second, the third and the fourth place with 0.9939, 0.9889 and 0.9495. Besides, we analyzed the performance of testing set.   For example, when $n=300$ and $\rho=0.6$, the values of AUC, sensitivity, specificity and accuracy of SPL-logsum are decreased to 0.9222 to 0.9310, 0.8750, 0.9000 and 0.8889, in which $\sigma$ is from 0.3 to 0.9. When the sample size $n$ increases, the  performances of all the four methods are improved. For example, when $\rho=0.2$ and $\sigma=0.3$, for testing set, the AUC and accuracy of  SPL-Logsum is increased by 11$\%$ and 10$\%$ with $n=200$ and $300$. When the $\rho$ decreases, the performance of four methods  are decreased with $\sigma=0.9$. For example, when $n=300$,the results of SPL-logsum are decreased from 0.9310, 0.8750, 0.9000 and 0.8889 to 0.9030, 0.9070, 0.8511 and 0.8778 from $\rho=0.6$ to $\rho=0.2$. In a word, our proposed SPL-logsum approach is superior to  $L_1$, $L_{1/2}$ and Logsum in the simulation dataset.

\subsection{Analyses of real data}
Three public QSAR datasets are got from website, including AID 651580, AID 743297 and AID 74362. We utilized random sampling to divide datasets into training datasets and testing dataset(70\% for training set and 30\% for testing set). A brief description of these datasets is shown in Table \ref{table3}-\ref{table4}.
\begin{table}[htbp]
\footnotesize
  \centering
  \caption{Three publicly available QSAR datasets
used in the experiments}
  \setlength{\tabcolsep}{1.5mm}{
    \begin{tabular}{llc}
    \toprule
    \textit{\textbf{Dataset  }} & \textit{\textbf{No.of Samples(class1/class2)}} & \multicolumn{1}{l}{\textit{\textbf{No.of descriptors}}} \\
    \midrule
    \textbf{AID:651580} & 1614(914 ACTIVE/700 INACTIVE) & 1642 \\
    \textbf{AID:743297} & 200(58 ACTIVE/ 142 INACTIVE) & 1588 \\
    \textbf{AID:743263} & 983(229 ACTIVE/ 754 INACTIVE) & 1612 \\
    \bottomrule
    \end{tabular}}%
  \label{table3}%
\end{table}%

\begin{table}[htbp]
\scriptsize
  \centering
  \caption{The detail information of three QSAR datasets used in the experiments}
    \setlength{\tabcolsep}{1.5mm}{
    \begin{tabular}{lll}
    \toprule
    \textit{\textbf{Dataset  }} & \textit{\textbf{No.of Traing(class1/class2)}} & \textit{\textbf{No.of Testing(class1/class2)}} \\
    \midrule
    \textbf{AID:651580} & 1130(634 ACTIVE/496 INACTIVE) & 484(280 ACTIVE/204 INACTIVE) \\
    \textbf{AID:743297} & 140(37 ACTIVE/103 INACTIVE) & 60(39 ACTIVE/21 INACTIVE) \\
    \textbf{AID:743263} & 689(152 ACTIVE/537 INACTIVE) & 294(77 ACTIVE/217 INACTIVE) \\
    \bottomrule
    \end{tabular}}%
  \label{table4}%
\end{table}%

\begin{table}[htbp]
\footnotesize
  \centering
  \caption{The results obtained by four methods.}
    \setlength{\tabcolsep}{1.6mm}{
    \begin{tabular}{cccccc}
    \toprule
    \multirow{2}[4]{*}{\textit{\textbf{Dataset}}} & \multirow{2}[4]{*}{\textit{\textbf{Methods }}} & \multicolumn{4}{c}{\textit{\textbf{Testing dataset}}} \\
\cmidrule{3-6}          &       & \textit{\textbf{AUC}} & \textit{\textbf{Sensitivity}} & \textit{\textbf{Specificity}} & \textit{\textbf{Accuracy}} \\
    \midrule
    \multirow{4}[2]{*}{\textit{\textbf{AID:651580}}} & \textbf{$L_1$} & 0.7063 & 0.6000 & 0.6800 & 0.6333 \\
          & \textbf{$L_{1/2}$} & 0.6697 & 0.7353 & 0.6923 & 0.7167 \\
          & \textbf{Logsum} & 0.8251 & 0.7429 & 0.8000 & 0.7667 \\
          & \textbf{SPL-Logsum} & \textit{\textbf{0.8583}} & \textit{\textbf{0.8000}} & \textit{\textbf{0.8000}} & \textit{\textbf{0.8000}} \\
    \midrule
    \multirow{4}[2]{*}{\textit{\textbf{AID:743297}}} & \textbf{$L_1$} & 0.7368 & 0.6667 & 0.7143 & 0.7000 \\
          & \textbf{$L_{1/2}$} & 0.7045 & 0.7619 & 0.7179 & 0.7100 \\
          & \textbf{Logsum} & 0.7765 & 0.7778 & 0.6905 & 0.7167 \\
          & \textbf{SPL-Logsum} & \textit{\textbf{0.7816}} & \textit{\textbf{0.7667}} & \textit{\textbf{0.7692}} & \textit{\textbf{0.7333}} \\
    \midrule
    \multirow{4}[2]{*}{\textit{\textbf{AID:743263}}} & \textbf{$L_1$} & 0.7132 & 0.6986 & 0.7344 & 0.7153 \\
          & \textbf{$L_{1/2}$} & 0.7009 & 0.7286 & 0.6716 & 0.7007 \\
          & \textbf{Logsum} & 0.6939 & 0.6883 & 0.7000 & 0.7034 \\
          & \textbf{SPL-Logsum} & \textit{\textbf{0.7552}} & \textit{\textbf{0.7143}} & \textit{\textbf{0.7667}} & \textit{\textbf{0.7372}} \\
    \bottomrule
    \end{tabular}}%
  \label{table5}%
\end{table}%
\begin{figure}[htbp]
    \centering
    \includegraphics[width=8cm]{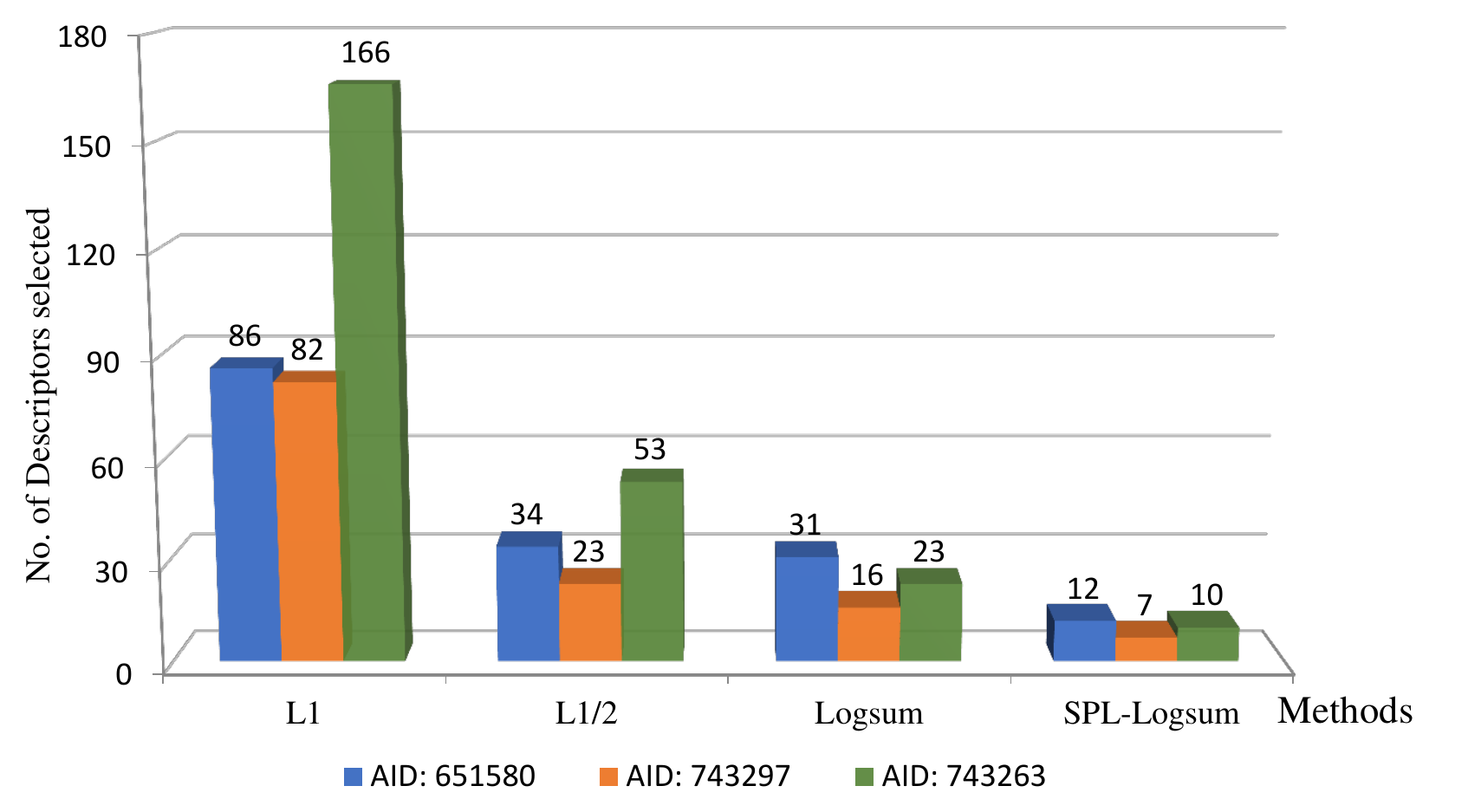}
    \caption{The number of descriptors got by $L_1$,$L_{1/2}$, Logsum and SPL-Logsum on different datasets}
    \label{figure4}
\end{figure}

Table \ref{table5} shows that the experiment results are got by $L_1$ , $L_{1/2}$, Logsum and SPL-Logsum for testing data. Our proposed SPL-Logsum is better than other methods in terms of AUC, sensitivity, specificity and accuracy. For example, from the view of dataset AID:743297, the AUC  obtained by our proposed  SPL-logsum is 0.7816 higher than 0.7368, 0.7045 and 0.7765 of $L_1$, $L_{1/2}$ and Logsum. Moreover, the accuracy of SPL-logsum is about increased by 17$\%$, 9$\%$, 4$\%$ of $L_1$, $L_{1/2}$ and Logsum for dataset AID:651580. Furthermore, our proposed SPL-Logsum method is more sparsity than other methods. For example, for dataset AID:743297 in Fig. \ref{figure4}, the number of selected descriptors of SPL-logsum is 10, ranks the first; then next is Logsum with 23; followed by $L_{1/2}$,constituting 53; finally, it comes from $L_1$ at 166, respectively.
\begin{table}[h]
\scriptsize
  \centering
  \caption{The 12 top-ranked meaningful and significant descriptors identified by $L_1$, $L_{1/2}$, Logsum and SPL-logsum from the AID: 651580 dataset.
(the common descriptors are emphasized in bold.)}
\setlength{\tabcolsep}{2mm}{
    \begin{tabular}{ccccc}
    \toprule
    \multirow{2}[4]{*}{\textbf{Rank}} & \multicolumn{4}{c}{\textbf{AID:651580}} \\
\cmidrule{2-5}          & $L_1$    & \textbf{$L_{1/2}$} & logsum & SPL-logsum \\
    \midrule
    \textbf{1} & L3i   & MLFER\_BH & MATS2e & \textit{\textbf{gmin}} \\
    \textbf{2} & L3u   & \textit{\textbf{MATS8m}} & ATSC6v & SHBint8 \\
    \textbf{3} & L3s   & maxHCsats & nHBint4 & GATS2s \\
    \textbf{4} & L3e   & SpMin8\_Bhe & \textit{\textbf{gmin}} & TDB6e \\
    \textbf{5} & maxtN & maxsssN & \textit{\textbf{ATSC3e}} & C1SP1 \\
    \textbf{6} & SssNH & mintsC & AATSC8v & MLFER\_A \\
    \textbf{7} & SpMAD\_Dzs & \textit{\textbf{ATSC3e}} & GATS8e & \textit{\textbf{MATS8m}} \\
    \textbf{8} & mintN & TDB5i & SHBint9 & minHBint4 \\
    \textbf{9} & \textit{\textbf{MATS8m}} & naaO  & mindssC & ATSC2i \\
    \textbf{10} & nHssNH & ATSC4i & ASP-3 & GATS3m \\
    \textbf{11} & RDF40s & SssssC & SsssCH & E2p \\
    \textbf{12} & RDF85s & AATSC4s & MATS1e & nHdCH2 \\
    \bottomrule
    \end{tabular}}%
  \label{table6}%
\end{table}%

\begin{table}[h]
\footnotesize
  \scriptsize
  \caption{The 7 top-ranked meaningful and significant descriptors identified by $L_1$, $L_{1/2}$,Logsum and SPL-Logsum from the AID: 743297 dataset.
(the common descriptors are emphasized in bold.)}
\setlength{\tabcolsep}{2mm}{
    \begin{tabular}{ccccc}
    \toprule
    \multirow{2}[4]{*}{\textbf{Rank}} & \multicolumn{4}{c}{\textbf{AID:743297}} \\
\cmidrule{2-5}          & $L_1$    & \textbf{$L_{1/2}$} & logsum & SPL-logsum \\
    \midrule
    \textbf{1} & SIC1  & \textit{\textbf{AATS3s}} & VCH-3 & nAcid \\
    \textbf{2} & BIC1  & ndsN  & ATSC5e & ATSC3s \\
    \textbf{3} & MATS7v & \textit{\textbf{RDF30m}} & \textit{\textbf{AATSC7v}} & AATSC5e \\
    \textbf{4} & \textit{\textbf{AATSC7v}} & ETA\_Shape\_P & nHAvin & SsssCH \\
    \textbf{5} & MLFER\_S & GATS4c & \textit{\textbf{AATS3s}} & minHCsatu \\
    \textbf{6} & \textit{\textbf{RDF30m}} & nsOm  & ATSC7i & minssNH \\
    \textbf{7} & RDF70p & ATSC4c & GATS2p & RDF100v \\
    \bottomrule
    \end{tabular}}%
  \label{table7}%
\end{table}%

\begin{table}[h]
\scriptsize
  \centering
  \caption{ The 10 top-ranked meaningful and significant descriptors identified by $L_1$, $L_{1/2}$,Logsum and SPL-Logsum from the AID: 743263 dataset.
(the common descriptors are emphasized in bold.)}
\setlength{\tabcolsep}{2mm}{
    \begin{tabular}{ccccc}
    \toprule
    \multirow{2}[4]{*}{\textbf{Rank}} & \multicolumn{4}{c}{\textbf{AID:743263}} \\
\cmidrule{2-5}          & $L_1$    & \textbf{$L_{1/2}$} & logsum & SPL-logsum \\
    \midrule
    \textbf{1} & maxsF & RDF130u & LipinskiFailures & AATS5v \\
    \textbf{2} & minsF & \textit{\textbf{nBondsD2}} & nF7Ring & AATS6s \\
    \textbf{3} & SRW9  & RDF25m & SCH-3 & ATSC2i \\
    \textbf{4} & SRW7  & nHssNH & SssNH & MATS3m \\
    \textbf{5} & MDEC-13 & minaaN & minHBint2 & \textit{\textbf{nBondsD2}} \\
    \textbf{6} & ETA\_dAlpha\_A & SpMin7\_Bhi & ATSC3p & nHBint3 \\
    \textbf{7} & minHCsats & nHdsCH & \textit{\textbf{RDF45s}} & minHAvin \\
    \textbf{8} & RDF95u & \textit{\textbf{RDF45s}} & ATSC1v & maxHBint10 \\
    \textbf{9} & TDB10m & \textit{\textbf{C3SP2}} & MDEO-11 & nFRing \\
    \textbf{10} & \textit{\textbf{C3SP2}} & RDF75v & AATSC8p & RDF20s \\
    \bottomrule
    \end{tabular}}%
  \label{table8}%
\end{table}%

\begin{table*}[h]
  \centering
  \caption{The detail information of descriptors obtained by SPL-logsum method.}
  \scalebox{0.94}{
    \begin{tabular}{llcc}
    \toprule
    \textit{\textbf{Descriptors}} & \textit{\textbf{Name}} & \textit{\textbf{P-value}} & \textit{\textbf{Class}} \\
    \midrule
    gmin  & Minimum E-State & 0.00050 & 2D \\
    TDB6e & 3D topological distance based autocorrelation - lag 6 / weighted by Sanderson electronegativities & 0.00368 & 3D \\
    GATS2s & Geary autocorrelation - lag 2 / weighted by I-state & 0.00576 & 2D \\
    C1SP1 & Triply bound carbon bound to one other carbon  & 0.00849 & 2D \\
    MLFER\_A & Overall or summation solute hydrogen bond acidity & 0.01383 & 2D \\
    SHBint8 & Sum of E-State descriptors of strength for potential hydrogen bonds of path length 8 & 0.01620 & 2D \\
    MATS8m & Moran autocorrelation - lag 8 / weighted by mass & 0.02395 & 2D \\
    E2p   & 2nd component accessibility directional WHIM index / weighted by relative polarizabilities & 0.03345 & 3D \\
    minHBint4 & Minimum E-State descriptors of strength for potential Hydrogen Bonds of path length 4 & 0.04056 & 2D \\
    nHdCH2 & Minimum atom-type H E-State: =CH2 & 0.04883 & 2D \\
    GATS3m & Geary autocorrelation - lag 3 / weighted by mass & 0.05929 & 2D \\
    ATSC2i & Centered Broto-Moreau autocorrelation - lag 2 / weighted by first ionization potential & 0.06968 & 2D \\
    nAcid & Number of acidic groups. The list of acidic groups is defined by these SMARTS "([O;H1]-[C,S,P]=O)" & 0.00569 & 2D \\
    ATSC3s & Centered Broto-Moreau autocorrelation - lag 3 / weighted by I-state & 0.00866 & 2D \\
    AATSC5e & Average centered Broto-Moreau autocorrelation - lag 5 / weighted by Sanderson electronegativities & 0.00210 & 2D \\
    SsssCH & Sum of atom-type E-State: $>$ CH- & 0.01260 & 2D \\
    minHCsatu & Minimum atom-type H E-State: H?on C sp3 bonded to unsaturated C & 0.04492 & 2D \\
    minssNH & Minimum atom-type E-State: -NH2-+ & 0.00025 & 2D \\
    RDF100v & Radial distribution function - 100 / weighted by relative van der Waals volumes & 0.01730 & 3D \\
    AATS5v & Average Broto-Moreau autocorrelation - lag 5 / weighted by van der Waals volumes & 0.08453 & 2D \\
    AATS6s & Average Broto-Moreau autocorrelation - lag 6 / weighted by I-state & 0.03047 & 2D \\
    ATSC2i & Centered Broto-Moreau autocorrelation - lag 2 / weighted by first ionization potential & 0.00536 & 2D \\
    MATS3m & Moran autocorrelation - lag 3 / weighted by mass & 0.00037 & 2D \\
    nBondsD2 & Total number of double bonds (excluding bonds to aromatic bonds) & 0.00059 & 2D \\
    nHBint3 & Count of E-State descriptors of strength for potential Hydrogen Bonds of path length 3 & 0.00041 & 2D \\
    minHAvin & Minimum atom-type H E-State: H on C vinyl bonded to C aromatic & 0.02756 & 2D \\
    maxHBint10 & Maximum E-State descriptors of strength for potential Hydrogen Bonds of path length 10 & 0.04263 & 2D \\
    nFRing & Number of fused rings & 0.03005 & 2D \\
    RDF20s & Radial distribution function - 020 / weighted by relative I-state & 0.05537 & 3D \\
    \bottomrule
    \end{tabular}}%
  \label{table9}%
\end{table*}%

Table \ref{table6},\ref{table7} and \ref{table8} show that the number of top-ranked informative molecular descriptors extracted by $L_{1}$, $L_{1/2}$ and SPL-logsum are 12, 7 and 10 based on the value of coefficients. Moveover, the common descriptors are emphasized in bold. Furthermore, as shown in Table \ref{table9}, the selected molecular descriptors got by our proposed SPL-logsum method are meaningful and significant in terms of p-value and almost belong to the class 2D. To sum up, our proposed SPL-Logsum is the effective technique for descriptors selection in real classification problem.

\section{Conclusion}
In the field of drug design and discovery, Only a few descriptors related to bioactivity are selected that can be for QSAR model of interest. Therefore, descriptor selection is an attractive method that reflect the bioactivity in QSAR modeling. In this paper, we proposed a novel descriptor selection using SPL via sparse LR with Logsum  penalty in QSAR classification. SPL can identify the easy and hard samples adaptively according to what the model has already learned and gradually add harder samples into training and Logsum regularization can select few meaningful and significant molecular descriptors simultaneously, respectively.

Both experimental results on artificial and three QSAR datasets demonstrate that our proposed SPL-Logsum method is superior to $L_1$, $L_{1/2}$ and Logsum. Therefore, our proposed method is the effective technique in both descriptor selection and prediction of biological activity.

In this paper,  SPL has capacity to learn from easy samples to hard samples. However, we ignore an important aspect in learning: diversity. We plan to incorporate this information into our proposed
method in our future work.

\appendices
\section{Abbreviations:}
    \begin{tabular}{ll}
    QSAR  & Quantitative structure-activity relationship \\
    QSRR  & Quantitative structure-(chromatographic) \\
    & retention relationships \\
    QSPR  & Quantitative structure-property relationship \\
    QSTR  & Quantitative structure-toxicity relationship \\
    MCP   & Maximum concave penalty \\
    SCAD  & Smoothly clipped absolute deviation \\
    $L_1$    & LASSO \\
    $L_{EN}$   & Elastic net \\
    LR    & Logistic regression  \\
    SPL   & Self-paced learning \\
    AUC   & Area under the curve \\
    \end{tabular}%



\bibliography{IEEEabrv,template}

\end{document}